\documentclass[a4paper,UKenglish,cleveref, autoref, thm-restate]{lipics-v2021}
\def\rnd#1{\num[round-mode=places,round-precision=3]{#1}}

\def\smallq{\textit{small}} 
\def\largeq{\textit{large}}
\usepackage{siunitx}
\usepackage{orcidlink}

\title{Evaluating the Ability of Large Language Models to Reason about Cardinal Directions} 

\titlerunning{LLM's abilities to reason about Cardinal Directions}

\author{Anthony G Cohn
}{School of Computing, University of Leeds, UK}{}{0000-0002-7652-8907}{AGC thanks the Turing’s Defence and Security programme through a partnership with the UK government in accordance with the framework agreement between GCHQ and The Alan Turing Institute, and for support provided by the Economic and Social Research Council (ESRC) under grant ES/W003473/1.}
\author{Robert E Blackwell
}{Alan Turing Institute, UK}{}{0000-0002-0554-8062}{}

\authorrunning{A G Cohn et al.} 

\Copyright{Anthony G Cohn and Robert E Blackwell} 



\category{} 

\relatedversion{} 

\supplement{}
\supplementdetails[subcategory={}, cite={}, swhid={}]{Dataset}{https://tinyurl.com/COSIT24-CDs}

\acknowledgements{
We thank the anonymous referees for their helpful comments.
This work was supported by the Fundamental Research priority area of The Alan Turing Institute.
We also thank Microsoft Research - Accelerating Foundation Models Research program, for the provision of Azure resources to access GPT which were used in the early stages of the work.
}

\nolinenumbers 

\EventEditors{B Adams, A Griffin, S Scheider and G McKenzie}
\EventNoEds{2}
\EventLongTitle{16th Conference on Spatial Information Theory (COSIT 2024)}
\EventShortTitle{COSIT 2024}
\EventAcronym{COSIT}
\EventYear{2024}
\EventDate{September 17--20, 2024}
\EventLocation{Quebec City, Canada}
\EventLogo{}
\SeriesVolume{?}
\ArticleNo{?}

\begin{document}

\maketitle

\begin{abstract}
We investigate the abilities of a representative set of Large language Models (LLMs) to reason about cardinal directions (CDs). To do so, we create two datasets: the first, co-created with ChatGPT, focuses largely on recall of world knowledge  about CDs; the second is  generated from a set of templates, comprehensively testing an LLM's ability to  determine the correct CD given a particular scenario.  The templates allow for a number of degrees of variation such as means of locomotion of the agent involved, and whether set in the first , second or third person. Even with a temperature setting of zero, Our experiments show that although LLMs are able to perform well in the simpler dataset, in the second more complex dataset no LLM is able to reliably determine the correct CD, even with a temperature setting of zero.
\end{abstract}
\section{Introduction}
Many claims (e.g. \cite{creswell2022faithful,huang2023reasoning,kojima2022large}) have been made since the emergence of Large Language Models (LLMs) as to their ability to reason. Spatial reasoning is of particular interest since not only does it underlie a  human's  ability to operate in the physical world, but also because LLMs are not embodied; 
so the question arises, have they nonetheless acquired an ability to reason about situations which might occur in the real world?  This is the question we address here. Spatial reasoning in general encompasses many aspects including topology, distance, and direction.  Here, we restrict ourselves to  reasoning about
cardinal directions (CDs). CDs are important for many reasons, e.g.: (i) successful navigation and route finding/following usually requires a fundamental understanding and ability to reason about CDs: CDs are crucial to success when using a map. Equally, giving directions often relies, at least in part, on the use of CDs. (ii) Understanding the geography of
an area
depends on understanding the relative CD of one location to another -- migration patterns, climate differences and economic variations are often underpinned by CDs. (iii) Weather patterns are often heavily influenced by the direction of the prevailing wind. (iv) CDs often play a critical role in cultural and historical contexts, e.g. the alignment of the pyramids to the four
CDs, or in certain languages -- e.g. the aboriginal language \emph{Guugu Yimithirr}  has no words for left or right, and  spatial information is mainly conveyed using CDs \cite{haviland1998guugu}.

In this paper we therefore analyse how well LLMs
can reason about cardinal and inter-cardinal directions.
We do this by automatically constructing a large set of
questions based on templates, for which the correct answer has been pre-determined, and testing each LLM's ability to answer the questions correctly. We also tested the LLMs on a small set of simpler questions, co-created with ChatGPT, in which recall of world knowledge is more prevalent.

\section{Related Work}
Despite the rapidly growing amount of research into LLMs and their capabilities there has been relatively little devoted specifically to spatial and/or geographic reasoning, and none which has  tested their ability to reason about CDs in the way we do here. Of the existing work we note benchmarks such as StepGame \cite{li2024advancing,shi2022stepgame} which aim to test an LLM's ability to correctly determine the spatial relationship between two objects, given the spatial relations between a larger set of objects, and between 1 and 10 reasoning steps are required to correctly determine the result;  the direction relationships are not exclusively  CDs, but also include, for example ``clock face directions'' (B is in the three o'clock direction from C). Not surprisingly performance deteriorates as the required number of steps increases. Performance increases markedly when the LLM is used to translate from the English specification to a logical representation and symbolic reasoning is used to compute the relationship. The SpartQA dataset \cite{mirzaee-etal-2021-spartqa} is also focused on assessing spatial reasoning, but does not contain any CDs.  The bAbI dataset \cite{weston2016towards} has one task which tests CDs understanding,  task 19,  which contains 1000 training and 1000 test questions: each instance contains 5 facts stating CDs between two objects, and then a question asking about the relation between two of them. Other work \cite{yamada2024evaluating} has investigated whether LLMs can acquire an understanding of a spatial environment from a turn-by-turn description of a route, with landmarks named at each turn; whilst the LLMs did perform reasonably well, the experiment did not involve any CDs, only left/right and up/down.

There are different kinds of spatial reasoning tasks which can be considered.  Relational composition is one of the most studied from a theoretical point of view.
A \emph{composition table} records the results for all combinations of relations in a particular spatial representation such as RCC; an investigation \cite{cohn2023evaluation} into ChatGPT's abilities to compute all RCC compositions found reasonable accuracy levels (reduced when  relations are anonymised); however RCC is a purely mereotopological calculus with no notion of direction embedded in its semantics.

Some LLMs have been built specifically for geo-applications, but these do not focus on reasoning about directions but rather aspects such as toponym recognition, e.g. \cite{Li2023GeoLMEL}.

\section{Experimental Design}
Whilst  testing compositional reasoning with CDs would be of interest, here we restrict ourselves to testing simpler reasoning abilities.
We created two question and answer sets which we refer to as \smallq{} and \largeq{}.
For \smallq{}
we used ChatGPT to co-create 100 simple questions where the answer is a CD \( \{north, south, east, west\} \). We edited the questions and corrected the answers where necessary. We changed the questions to ensure equal class representation amongst the four answers. Example questions are:
\begin{itemize}
\item \textit{You are watching the sun set. Which direction are you facing?}
\item \textit{If the South Pole is behind you, which direction are you facing?}
\end{itemize}
We use this dataset to give an overall assessment of LLM performance in real world scenarios requiring directional common sense spatial reasoning and common sense spatial knowledge.

It would be impractical to generate a substantial question set manually and so we used an automated, template driven approach for
\largeq{}.
We wanted to investigate the ability of LLMs to reason about CDs in the context of a simple scenario involving locomotion along or around a geographical feature as this is a test of an LLM in a realistic situation. Based on some informal experimentation using GPT and ChatGPT with a selection of questions and noting a lack of accuracy,  we chose six question templates to test LLM performance more comprehensively:
\begin{itemize}

\item \textit{You are walking [south] along the [east] shore of a lake; in which direction is the lake?} (Template T1).
\item \textit{You are walking [south] along the [east] shore of a lake and then turn around to head back in the direction you came from, in which direction is the lake?} (Template T2).
\item  \textit{You are walking [south] along the middle of the [east] side of a park; in which direction is the bandstand located in the centre of the park?} (Template T3).
\item  \textit{You are walking [east] along the [south] side of a road which runs [east to west]. In which direction is the road?} (Template T4).
\item  \textit{You are walking [south] along the [east] shore of the island. In which direction is the sea?} (Template T5).
\item  \textit{You are walking [south] along the [east] shore of an island and then turn around to head back in the direction you came from, in which direction is the sea?} (Template T6).
\end{itemize}

We then exhaustively generated all forms of these questions for all cardinal and inter-cardinal directions and ten different locomotion types
\emph{\{cycling,
driving,
hiking,
jogging,
perambulating,
racing,
riding,
running,
unicycling,
walking\}}. Note that in each template, once one of the directions (between ``[ ]'') is fixed, then there are only two possibilities for the second direction.  Following earlier evidence \cite{leyton-brown-slides} that an LLM's performance can vary depending on which person a question is phrased as using, we also generated questions in the first-person (\emph{I am}), first-person plural (\emph{We are}), second-person (\emph{You are}), third-person singular (\emph{He is} and \emph{She is}), and third-person plural (\emph{They are}) forms. This gives us 6 questions $\times$ 10 forms of locomotion $\times$ 6 person forms $\times$ 8 directions $\times$ 2 directions-variations = 5760 questions.

Previous work suggests that models with less than about 40B parameters perform poorly at reasoning \cite{leyton-brown-slides}. We therefore favour larger models, testing those listed in Table~\ref{tab:models}.

\begin{table}[htb]
\footnotesize
\centering
\begin{tabular}{|c|c|c|c|c|}
\hline
\textbf{API} & \textbf{Model} & \textbf{Released} & \textbf{
Num. params} & \textbf{Window} \\ \hline
Anthropic Claude& claude-3-opus-20240229 & Feb 2024 & 137B & 200,000  \\
Google Vertex & gemini-10-pro & Dec 2023 & 1.6T &  32,000   \\
 & gemini-15-pro-preview-0409 & Apr 2024 & $\gg$3.5T & 128,000   \\
OpenAI & gpt-3.5-turbo-0613 \cite{brown2020language}  & Jun 2023 & 175B & 4,096   \\
  & gpt-35-turbo-1106 & Nov 2023 & 175B & 16,385  \\
  & gpt-3.5-turbo-0125 & Jan 2024 & 175B & 16,385  \\
& gpt-4-0613 \cite{openai2024gpt4} & Jun 2023 & 1.76T & 8,192   \\
  & gpt-4-turbo-2024-04-09 & Apr 2024 & 1.76T & 128,000   \\ \hline
\end{tabular}
\caption{LLMs tested.
in our experiments.
Window is the context window size in tokens.}
\label{tab:models}
\end{table}

\subsection{Prompting}

Zero-shot prompting is when a model is given a question without any
examples to help guide the answer.
The model must then attempt to answer the question based solely on its general pre-training. As a system prompt, we use  "You are a helpful assistant. I will give you a question about directions. The answer is either north, south, east, west, north-east, north-west, south-east or south-west. Please only reply with the answer. No yapping.". We then present each question in a new chat.  We include "No yapping." since that has been reported (\url{https://tinyurl.com/no-yapping} as being beneficial in persuading an LLM to be brief in its response.

We set temperature = 0 for each model to try to achieve deterministic answers. Temperature is a parameter that affects the randomness or variability of the responses generated by a language model and helps to control the predictability of the the model's output but even a 0 temperature does not guarantee reproducible, deterministic behaviour. To explore the effect of temperature on accuracy, we take our best performing model on the \emph{large} dataset and vary temperature \(t\) , \(0 \leq t \leq 2\).

We use case-insensitive string comparison and remove spurious punctuation and white space before comparing answers; e.g. we regard "`North East'." and "north-east" as equivalent.
Our prompts ask for cardinal or inter-cardinal direction answers only: we count answers such as "The lake is to the west" as correct if the intended answer is "west", but we note instances where answers do not strictly meet the rubric. We assess performance using accuracy, and report variability using the standard error of the mean.

\section{Results}

\begin{figure}[ht!]
    \centering
    \includegraphics[page=1, width=0.98\textwidth]{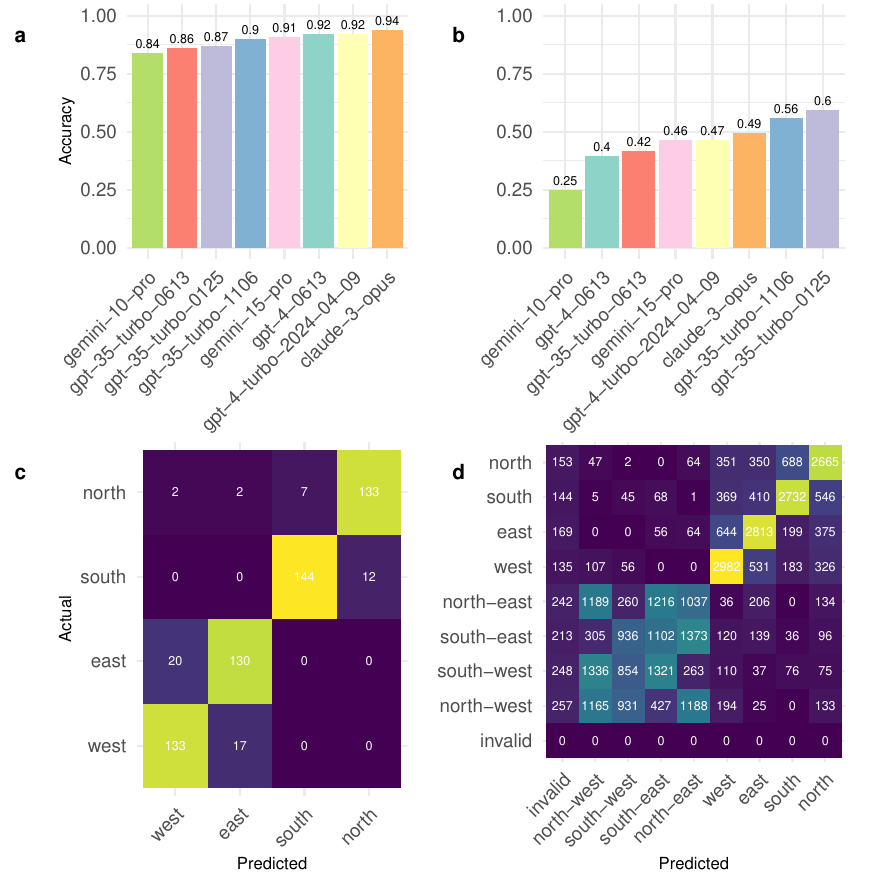}
    \caption{
    (a) and (c) show accuracy by LLM and confusion matrix respectively, for \smallq{} question set; (b) and (d) for  \largeq{}.
    Answers that cannot be interpreted as a CD or an inter-CD are considered invalid.
    To avoid bias from three gpt-35-turbo models, the confusion matrices exclude gpt-35-turbo-0613 and gpt-35-turbo-1106 but include gpt-35-turbo-0125. }
    \label{fig:summary}
\end{figure}

All models tested showed an accuracy of $> 0.8$ for \smallq{}  (Fig.~\ref{fig:summary}a). Where confusion occurred, it was mostly \textit{north} confused with \textit{south}, and \textit{east} confused with \textit{west} (Fig.~\ref{fig:summary}c). In one case a model ignored the rubric: gemini-1.0-pro  answered the question \textit{"On a hike, a duck pond is to your north and the nearest town is south. What direction is the pond from the town?"} with \textit{"The pond is north of the town"}, which is correct but not a one word answer. Of the 100 questions, 77 were always correctly answered.
Only one question was always answered incorrectly:
\textit{In a stadium with a north-facing entrance, if the VIP section is on the left side, which direction would it be in? (east)}; all answered \textit{west}.

Model accuracy was worse for  \largeq{}, (the more complex dataset), with the best performing model (gpt-35-turbo-0125) achieving only \rnd{0.5951389}. Of the 5760 questions, only 294 (5.10\%) were correctly answered by all the models. 628 questions (10.90\%) were not answered correctly by any  of the models.
Of those questions not answered correctly by any of the models, 368 (58.60\%) were T4 questions (suggesting roads running from one direction to another are a cause of confusion), 129 (20.54\%) were T6 and 98 (15.61\%) were T2 (suggesting that turning backwards is a difficulty).
The rubric was not followed for 1762 of the 46080 answers (3.82\%) from the eight models. gpt-4-turbo-2024-04-09 failed to follow the rubric on 618 questions (10.73\%). gpt-4-0613 failed to the follow the rubric on 608 questions (10.56\%)
claude-3-opus failed to follow the rubric on 207 questions (3.59\%). gemini-10-pro failed to follow the rubric on 190 questions (3.30\%). All other models followed the rubric on more than 98\% of questions, with gpt-35-turbo-0125 failing to follow the rubric on only one question. Only gemini-10-pro gave correct answers when not following the rubric (33 such answers, 17\%).

\begin{figure}[ht!]
    \centering
    \includegraphics[page=1, width=.98\textwidth]{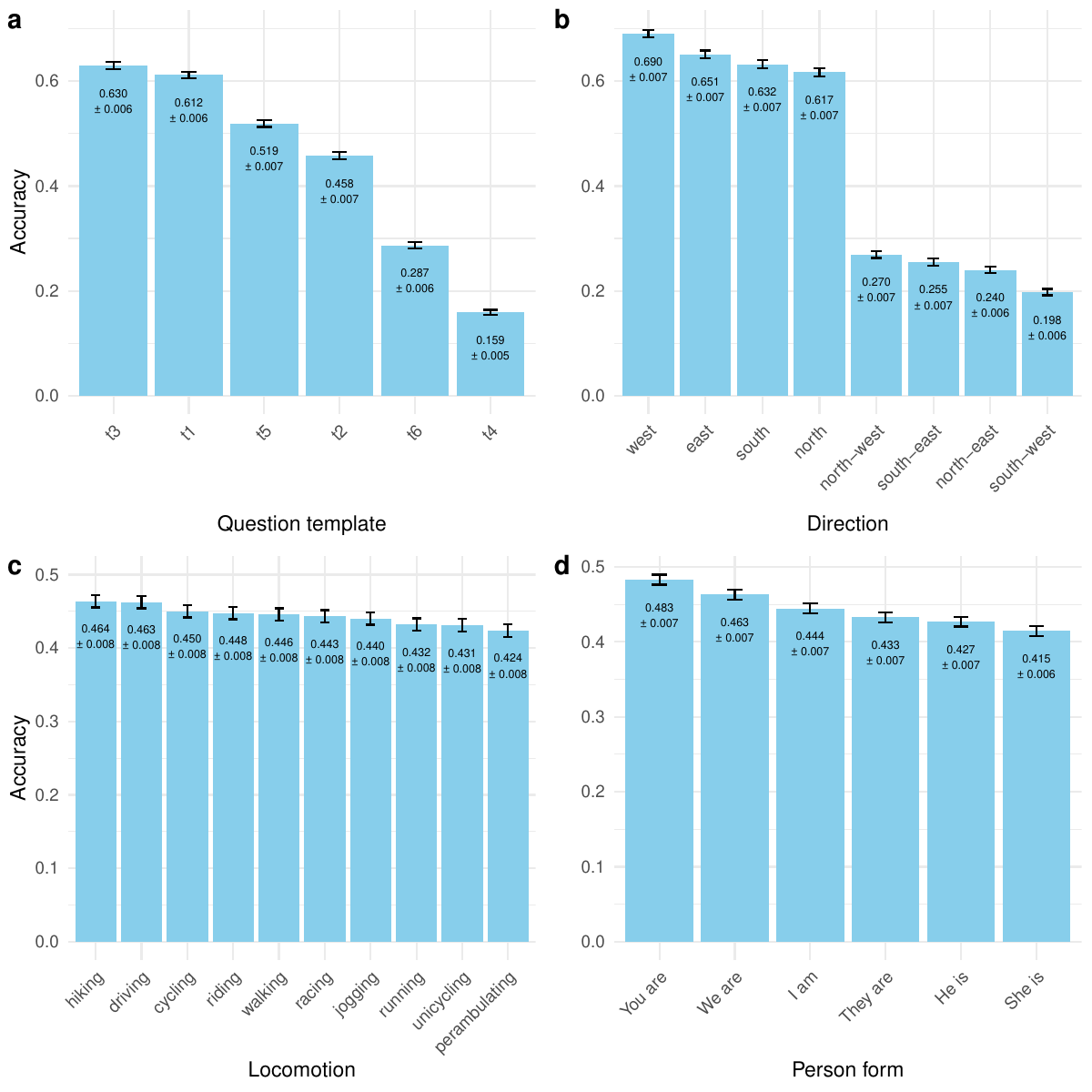}
    \caption{Accuracy by (a) question template, (b) direction, (c) locomotion and (d) person form for \largeq{}. To avoid bias from using three gpt-35-turbo models, we exclude gpt-35-turbo-0613 and gpt-35-turbo-1106 but include gpt-35-turbo-0125.}
    \label{fig:accuracy_by}
\end{figure}

1595 (90.52\%) of the answers where the rubric was not followed were answers to T4 questions, and the answer given was one of `north-south', `east-west',`south-east to north-west',`north-east to south-west',`south-west to north-east',`north-west to south-east',`south to north',`north to south',`west to east',`west-east',`east to west' (further suggesting roads running from one direction to another are a cause of confusion, Fig. \ref{fig:accuracy_by}a).

Models achieve higher accuracy on cardinal directions than inter-cardinal directions (Fig. \ref{fig:accuracy_by}b).
We found little difference in accuracy amongst the various forms of locomotion (Fig. \ref{fig:accuracy_by}c). These minor differences may be due to the incidence of the words in the training data of the models.
Second person prompts
(\emph{You are}) have the highest accuracy, followed by first  and then third person (Fig. \ref{fig:accuracy_by}d).

\begin{figure}[ht!]
    \centering
    \includegraphics[page=1, width=.98\textwidth]{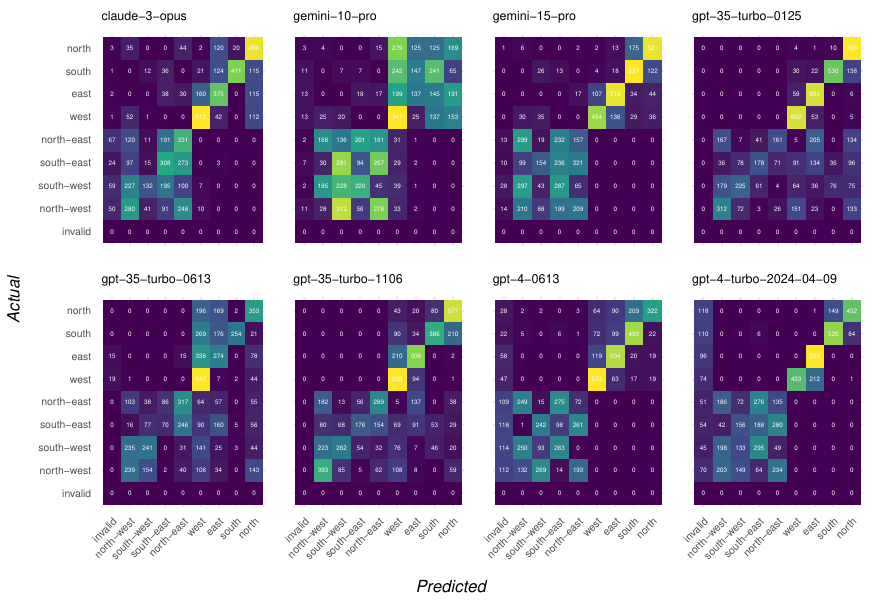}
    \caption{Confusion matrices for each of the LLMs used to test \largeq{}. Answers that cannot be interpreted as CD or inter-CD are considered invalid.}
    \label{fig:confusion}
\end{figure}


Fig.~\ref{fig:confusion} gives a  breakdown of confusion for each LLM. The upper right quadrant of each matrix gives the performance for the 4 main CDs and it can be seen that in general (except for gemini-10-pro and to a lesser extent gpt-35-turbo-0613) all models perform well here -- it is the inter-CD relations which cause most confusion. Surprisingly, there is asymmetry in the north/south and east/west confusion. For example, gpt-35-turbo-0125 predicted north when the answer was south on 138 occasions but predicted south when the answer was north on only ten occasions. gpt-4-turbo-2024-4-09 predicted east when the answer was west on 212 occasions but never predicted west when the answer was east. But this bias towards north and east is not universal -- some of the models have a reverse bias. We do not have a good explanation for this unexpected asymmetry.

\begin{figure}[ht!]
    \centering
    \includegraphics[page=1, width=.5\textwidth]{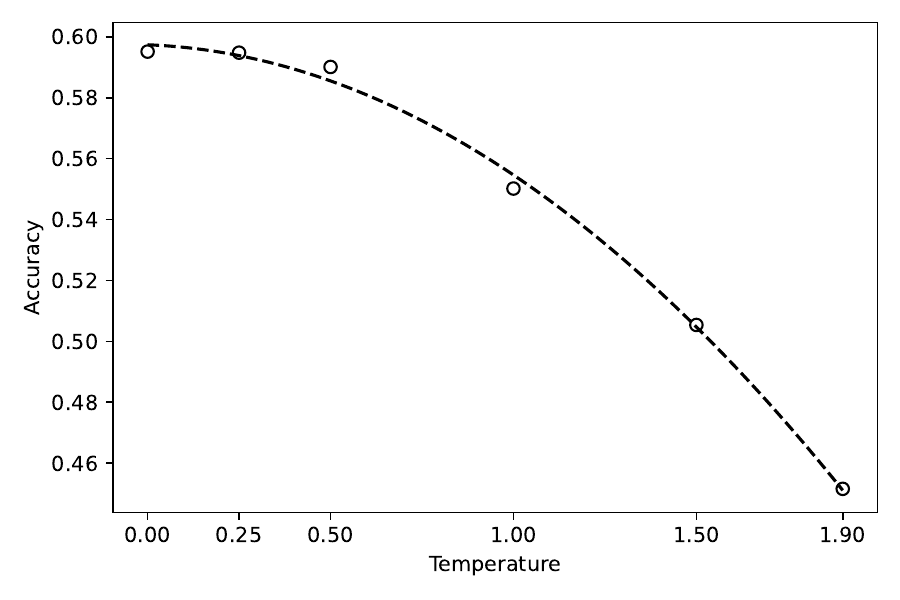}
    \caption{Accuracy by temperature for gpt-35-turbo-0125 applied to \largeq{}.}
    \label{fig:temperature}
\end{figure}

Fig.~\ref{fig:temperature} Shows accuracy by temperature for gpt-35-turbo-0125 applied to
\largeq{}. When temperature increases, accuracy decreases. This can be explained by more random next token prediction in the model.  As temperature, \(t
\to 2.0\) the number of errors from the model (\emph{HTPP 500 - The server had an error while processing your request. Sorry about that!}) also increases,
requiring repeated retries before obtaining any answer and making \(t = 2.0\)
impractical.

\section{Discussion and Conclusions}

None of the models tested is able to reliably reason about cardinal direction, however all models show some capacity for spatial reasoning.  A model that randomly selects cardinal or inter-cardinal direction answers to an MCQ would have an accuracy of 0.125, but even the worst performing model (gemini-10-pro) achieved an accuracy of 0.25 on  \largeq{}.

All models showed higher accuracy on \smallq{}  compared to  \largeq{}, though this would be expected given that \smallq{} only has four possible answers whilst  \largeq{} has eight. The questions in \largeq{} arguably all require reasoning and a model-based approach to solving, unlike  \smallq{}. It is unclear if LLMs produce the answer by reasoning or by recalling memorized information \cite{hou2023towards} -- \largeq{}  requires more reasoning,  \smallq{} relatively more factual recall.

Given the results here, LLMs appear to perform better at factual recall than spatial reasoning. Of the questions in  \smallq{}, 16 can be reasonably regarded as only requiring world knowledge, and 84 as requiring simple reasoning. All the LLMs answered the world knowledge questions correctly. Of all the answers to the simple reasoning questions, 87\% were correct.

Unlike results generally reported in the literature (e.g. \cite{raman2024rationality}) comparing GPT-35-Turbo and GPT-4 performance, for  \largeq{}, the OpenAI interface to GPT-35-turbo has the highest accuracy.  Moreover the latest GPT-35-Turbo beats the latest GPT-4-Turbo and the latter was released more recently than the former.  We do not currently  have an explanation for this. One author had experimented briefly with T1-T3 on ChatGPT-4, but never provided feedback, so contamination seems unlikely. The same author has also given talks at several venues using T1-T3 as examples, and it is conceivable that this might have reached OpenAI who may have improved their model as a result, but this does not explain why GPT-35-Turbo is better than GPT-4-Turbo, particularly when the release date of the former is before that of the latter.


The development of LLMs is progressing rapidly (though many believe they will never achieve AGI, let alone ever achieve reliable reasoning abilities, at least without a neuro-symbolic component):
the Open AI GPT-35-turbo model was updated twice in seven months. Using  \largeq{}  as a benchmark, we observed a 43\% increase in performance between the Open AI GPT-35-turbo 0613 and 0125 versions. However, any evaluation such as this can only ever be a snapshot evaluation, so we hesitate to draw conclusions in general as to which LLM (family) is better than another.

We also tested  Microsoft Azure API access to  gpt-35-turbo and gpt-4. Although the Microsoft Azure API is designed to include additional guardrails, compliance and data governance certification and enterprise support, we found that accuracy was similar to the Open AI API models gpt-35-0613 and gpt-4-0613 respectively. The Azure documentation does not specify which OpenAI model their models exploit.





Possibilities for future work include: (1) Improving the question design; there are minor flaws in our current questions, e.g. differing punctuation in T1-T6, and a potential ambiguity in T4 (since a road is a linear object it might have been clearer to add "from $\langle agent \rangle$" to make it clear that the question is not relating to the orientation of the road. (2) Exploration and/or  development of prompting strategies~\cite{bhandari2024survey} to improve performance -- either using general methods such as \emph{chain of thought} or \emph{tree of thoughts}, or spatial-specific ones such  as \emph{Visualization-of-Thought}\cite{wu2024visualization}. (3) Other LLMs could be evaluated, or existing ones fine-tuned for the tasks under consideration. (4) Extend the variety of experiments to create a comprehensive benchmark for evaluating reasoning about directions -- in this paper we have deliberately only considered questions whose answer is a CD, but a more comprehensive dataset would also consider other directions (left, right, behind, above...).  (5) Building a comprehensive benchmark for other aspects of spatial reasoning (e.g. topological, distance) and combinations of these; ideally these would be generated programmatically (cf \cite{li2024ijcai}). (6) Consider situations with more than two objects of interest, so that, e.g. compositional reasoning can be tested, and also reasoning about trajectories (cf \cite{weston2016towards}).

\section*{Data Access Statement}
The data associated with this paper are available from  the following github repository: \url{https://tinyurl.com/COSIT24-CDs}.

\section*{Contribution Statement}
AC conceived the original idea for the \emph{large} dataset and RB subsequently the \emph{small} dataset. RB implemented the benchmark in consultation with AC, and performed the evaluations. Both authors wrote the original draft of the paper and contributed to the subsequent drafts.

\bibliography{lipics-v2021-sample-article}

\end{document}